\documentclass{article} 

\PassOptionsToPackage{table}{xcolor}

\usepackage{iclr2027_conference,times}

\usepackage{amsmath,amsfonts,bm}

\def\eqref#1{equation~\ref{#1}}

\def\1{\bm{1}}

\DeclareMathAlphabet{\mathsfit}{\encodingdefault}{\sfdefault}{m}{sl}
\SetMathAlphabet{\mathsfit}{bold}{\encodingdefault}{\sfdefault}{bx}{n}

\usepackage{amssymb}
\usepackage{booktabs}
\usepackage{graphicx}
\usepackage{array}
\usepackage{multirow}
\usepackage{xcolor}          
\usepackage{caption}
\usepackage{subcaption}
\usepackage{enumitem}
\usepackage{microtype}
\usepackage{algorithm}
\usepackage{algpseudocode}

\usepackage{hyperref}
\usepackage{url}
\hypersetup{
  hidelinks,
  pdfauthor={Zhangquan Chen, Yaoxin Niu, Xiang An, Mingze Sun, Zhumei Wang, Chih-Ting Liao, Hongkun Cao, Ruqi Huang},
  pdftitle={HaPRL: Human-Anchored Process Reinforcement Learning for Visual Search Agent}
}

\definecolor{basegrey}{RGB}{240,241,243}
\definecolor{gaingreen}{RGB}{22,110,54}
\definecolor{lossred}{RGB}{178,38,30}
\definecolor{neutgrey}{RGB}{110,116,124}

\newcommand{\gain}[1]{{\color{gaingreen}\small #1}}
\newcommand{\loss}[1]{{\color{lossred}\small #1}}
\newcommand{\neut}[1]{{\color{neutgrey}\small #1}}

\newcommand{\method}{HaPRL}
\newcommand{\orl}{Outcome-RL}

\newcommand{\rout}{r_{\mathrm{out}}}
\newcommand{\rproc}{r_{\mathrm{proc}}}

\iclrfinalcopy
\begin{document}
\fancyhead{}
\renewcommand{\headrulewidth}{0pt}
\fancyfoot[L]{\ifnum\value{page}=1\footnotesize Preprint.\fi}

\noindent\rule{\linewidth}{1.5pt}
\begin{center}
{\fontsize{16}{19}\selectfont\bfseries
HaPRL: Human-Anchored Process\\
Reinforcement Learning for Visual Search Agent\par}
\vspace{0.10in}
\rule{\linewidth}{0.5pt}\par
\vspace{0.11in}
{\normalsize
\textbf{Zhangquan Chen}$^{1,\dagger}$ \quad \textbf{Yaoxin Niu}$^{1,2,\dagger}$ \quad
\textbf{Xiang An}$^{3}$ \quad \textbf{Mingze Sun}$^{1}$\\[0.35em]
\textbf{Zhumei Wang}$^{4}$ \quad \textbf{Chih-Ting Liao}$^{5}$ \quad
\textbf{Hongkun Cao}$^{2}$ \quad \textbf{Ruqi Huang}$^{1,*}$\par}
\vspace{0.11in}
{\small
$^{1}$ Tsinghua University \quad $^{2}$ Peng Cheng Laboratory \quad $^{3}$ LMMs-Lab\\
$^{4}$ Beijing Institute of Technology \quad $^{5}$ University of New South Wales\\[0.45em]
$^{\dagger}$ Equal contribution. \quad $^{*}$ Corresponding author.\\[0.35em]
Code: \href{https://github.com/zhangquanchen/HAPRL}{\textcolor{blue}{github.com/zhangquanchen/HAPRL}}\par}
\end{center}
\vspace{0.02in}

\begin{minipage}{\linewidth}
\centering
\includegraphics[width=0.90\linewidth,trim=0 0 56bp 0,clip]{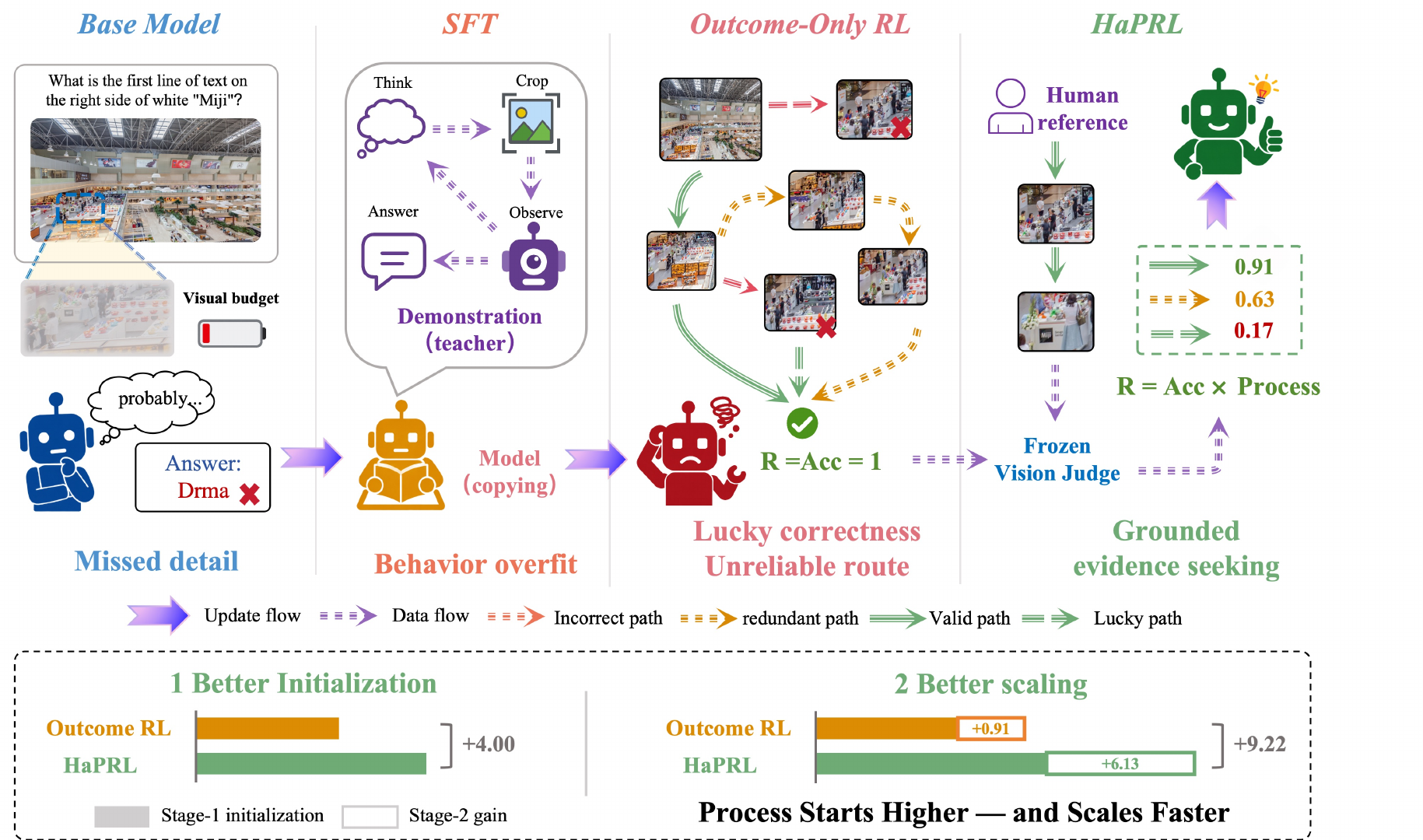}
\captionof{figure}{\textbf{Human-anchored process supervision.} Unlike outcome-only RL, HaPRL uses human traces to distinguish grounded from lucky correct searches. This improves initialization by \textbf{4.00} points and yields a \textbf{6.7×} larger gain in the same subsequent outcome-based stage.}
\label{fig:teaser}
\end{minipage}

\begin{abstract}

Multi-turn visual search agents answer questions about high-resolution images by iteratively deciding where to look.
Reinforcement learning for these agents rewards only the final answer, leaving the search process unsupervised.
Consequently, faulty routes in which the reasoning process is erroneous yet the final result is correct arise frequently, which \emph{in turn leads to ineffective training, i.e., scaling along the wrong paths.} 
In this paper, we introduce \method{}, the first framework to \emph{reinforce the search process with human search behavior}.
We first build an annotation platform and collect $1K+$ human-annotated data with fine-grained behavioral signals. During training, a carefully designed judge scores each rollout with task-adaptive weights, anchored on the distilled trace of how a human annotator actually searched the same image. 
Extensive experiments show that \method{} consistently outperforms outcome-based RL, and early-stage process supervision yields $6.7\times$ more improvement in subsequent outcome-based scaling. Our results also demonstrate the importance of aligning model behavior with human process annotation signals, which offer new insight into the training of foundation models.

\end{abstract}

\section{Introduction}
\label{sec:intro}

High-resolution visual reasoning often depends on evidence that occupies only a small fraction of an image. Human vision handles this constraint through sequential attention, turning a series of fixations into an efficient search process \citep{treisman1980feature,itti2001computational}. Multi-turn visual agents follow this principle by interleaving reasoning with crop-and-zoom tool calls, i.e., they choose a region, inspect the resulting view, and decide where to look next \citep{wu2024vstar,shao2024visualcot,hu2024visualsketchpad,zheng2025deepeyes,lai2025minio3, chen2025sifthinker, chen2025visrl}. A capable agent should reach the correct answer through a search trajectory that acquires the supporting evidence.

Training a capable visual search agent through supervised fine-tuning alone amounts to off-policy behavior cloning, which limits generalization beyond the demonstration distribution (Figure~\ref{fig:teaser}). Many methods therefore apply reinforcement learning after a cold start to unlock the agent's potential. Existing visual-search agents commonly adopt outcome-based rewards that score only the final answer \citep{zheng2025deepeyes,su2025pixelreasoner,lai2025minio3,liu2025visualarft}. 
Such outcome-based rewards are attractive because correctness is inexpensive to verify \citep{shao2024deepseekmath,deepseek2025r1,lambert2025tulu3}. However, \emph{human annotators also produce fine-grained behavioral signals through their mouse interactions during the search process}, such as movement trajectories, cursor velocity, dwell duration, etc.. \emph{These signals carry rich information about search quality but are entirely discarded} by outcome-based training. 
As a result, a correct answer reached by an incorrect search hacks full reward. On multiple-choice and short-answer tasks, for example, an agent may inspect irrelevant regions, miss the target evidence, and still arrive at the right answer through random guessing. Rewarding such trajectories \emph{reinforces faulty routes and causes optimization to scale along wrong search paths} (Figure~\ref{fig:teaser}).

More fundamentally, outcome-based GRPO~\citep{shao2024deepseekmath} standardizes rewards within each sampled group and provides no mechanism to rank correct trajectories by search quality. That is, a trajectory that localizes the sign, zooms in, and reads it receives the same credit as one that crops empty regions and guesses. We verify this on VisualProbe Hard, where outcome-based RL raises accuracy by $28.6\%$ while the judged search quality of its correct answers \emph{drops} by $8.4\%$ (Section~\ref{subsec:process_quality}), i.e., \emph{the agent answers more questions correctly yet searches worse doing so}. 

Process supervision addresses analogous failures in mathematical reasoning by evaluating intermediate steps \citep{uesato2022solving,lightman2024verify,wang2024mathshepherd,luo2024omegaprm}, but \emph{perceptual search does not decompose into self-contained steps} that can be labeled correct or incorrect. The value of a crop depends on the question, the regions already inspected, and whether the action makes progress toward the target. A useful process signal must therefore evaluate the trajectory as a whole rather than classify isolated coordinates. Human search behavior provides the missing reference, recording which fine-grained regions attracted attention, how the view was refined, and where the evidence was ultimately found. Addressing this requires solving two challenges: \textbf{G1) Human process signal acquisition}: systematically capture the fine-grained behavioral signals that human annotators produce during visual search; and \textbf{G2) Dense process alignment}: align agent training with these human signals to provide graded supervision.

For (G1), we build a browser-based annotation platform and collect human search traces of $1{,}104$ visual-search questions. The platform logs timestamped pointer movements, hovers, and zooms over the full-resolution image, preserving how annotators attend to and refine fine-grained regions. Each interaction stream is distilled into a compact round-by-round trace of the inspected regions and discovered evidence. \emph{These traces serve as instance-specific anchors that reveal where useful evidence lies and how much visual refinement the question requires.} For (G2), we introduce \method{}, a training paradigm that aligns visual-search agents with these human process signals at the trajectory level. A frozen vision-language judge evaluates each rollout along five dimensions with task-adaptive weights, anchored on the human-annotated trace of the same image. The resulting process score is multiplied by binary answer correctness, so incorrect answers receive no credit while correct trajectories are ranked by search quality.

Our contributions can be summarized as follows.
\begin{itemize}[leftmargin=1.4em,itemsep=1.5pt,topsep=2.5pt]
\item We develop a platform and collect $1{,}104$ human-annotated visual-search examples with process-level information, distilled into round-by-round traces of human attention and evidence acquisition.
\item We introduce \method{}, the first training paradigm that aligns visual-search agents with human process signals. The answer-gated, task-adaptive reward ranks trajectories anchored on traces.
\item Extensive experiments across four backbones and six benchmarks show that \method{} consistently outperforms outcome-based RL. Moreover, process annotation scales favorably, i.e., \emph{early-stage process supervision provides an initialization from which a subsequent outcome-based stage yields \textbf{6.7×} (6.13 vs. 0.91) more improvement than the same stage without it.}
\item Our results highlight the importance of aligning agent behavior with human process annotation signals, which offers practical insights for future efforts beyond visual search.
\end{itemize}

\section{Related Work}
\label{sec:related}

\paragraph{Visual Search Agents.} Visual search enables multimodal large language models (MLLMs) to inspect selected regions instead of relying on a single global view. V$^{\ast}$ combines LLM-guided search with multimodal reasoning to locate small targets in high-resolution images \citep{wu2024vstar}, while Visual Sketchpad equips models with drawing and specialist vision tools for intermediate visual reasoning \citep{hu2024visualsketchpad}. Recent work learns such interactions through post-training. DeepEyes develops active perception with end-to-end reinforcement learning \citep{zheng2025deepeyes}, and Pixel Reasoner combines instruction tuning with curiosity-driven reinforcement learning over visual operations \citep{su2025pixelreasoner}. Chain-of-Focus learns adaptive search and zooming through SFT followed by outcome- and format-based reinforcement learning \citep{zhang2025chainoffocus}, while Mini-o3 scales visual-search trajectories through diverse cold-start data and over-turn masking \citep{lai2025minio3}. Despite different tools and training recipes, \emph{these methods rely primarily on terminal correctness or hand-designed auxiliary signals}. We instead use human search traces as instance-specific references for evaluating complete evidence-acquisition trajectories.

\paragraph{Process Supervision for Reasoning.} Process supervision evaluates intermediate reasoning rather than relying exclusively on terminal outcomes. In mathematical reasoning, process-based feedback reduces reasoning errors among final-answer-correct solutions \citep{uesato2022solving}, and PRM800K scales human step-level feedback to the MATH dataset \citep{lightman2024verify}. Math-Shepherd and OmegaPRM reduce annotation costs by constructing process supervision automatically through repeated completions and tree search \citep{wang2024mathshepherd,luo2024omegaprm}. Multimodal PRMs extend step-level verification to image-conditioned mathematical reasoning, i.e. VisualPRM and MM-PRM score candidate derivations for Best-of-$N$ inference \citep{wang2025visualprm,du2025mmprm}, whereas URSA also incorporates process rewards into online policy optimization \citep{luo2025ursa}. These methods assess textual reasoning steps whose correctness can be evaluated individually. \emph{Visual search differs because every perceptual action changes the evidence available to subsequent decisions.} Accordingly, we evaluate the complete search trajectory against human behavior and use this trajectory-level signal for reinforcement learning.

\paragraph{Rubric-Based Reinforcement Learning.} LLM judges provide scalable model-based evaluation \citep{zheng2023judging, chen20264dthinker}, and recent work converts structured criteria into reinforcement signals. Rubrics as Rewards uses instance-specific rubrics for on-policy training in medical and scientific domains \citep{gunjal2025rubrics}. Reinforcement Learning from Checklist Feedback extracts instruction-specific criteria and aggregates judgments from language models and specialized verifiers \citep{viswanathan2025checklists}, while Rubric Anchors scales open-ended alignment with a large collection of human- and model-authored rubrics \citep{huang2025rubricanchors}. \emph{These methods primarily evaluate final responses when a single verifiable outcome is unavailable.} In visual search, the final answer is verifiable but does not reveal whether the agent acquired the necessary evidence. \method{} therefore applies a task-adaptive rubric to the search process, anchors the evaluation on a matched human trace, and gates the process score by answer correctness. 

\section{Method}
\label{sec:method}

\paragraph{Method Overview.} We study a visual-search policy $\pi_\theta$ that answers a question $q$ about a high-resolution image $I_0$ by interleaving language reasoning with crop actions, following the tool-use protocol of Mini-o3 \citep{lai2025minio3}. At round $t$, the policy emits a reasoning segment $h_t$ and either terminates with an answer $a$ or invokes $c_t=(v_t,b_t)$, where $v_t$ identifies a previously observed view and $b_t$ specifies a normalized bounding box. The environment executes the crop and returns the resulting observation $o_t$. A trajectory containing $T$ crop actions is
\begin{equation}
\tau=(h_1,c_1,o_1,\ldots,h_T,c_T,o_T,h_{T+1},a).
\label{eq:trajectory}
\end{equation}

The central idea of \method{} is to \emph{use human search behavior as a reward anchor rather than a policy demonstration.} As shown in Figure~\ref{fig:method_pipeline}, the policy samples multiple trajectories for each image-question pair. A frozen vision-language judge compares each search process with the matched human trace and produces a task-adaptive process score, while a separate answer judge evaluates terminal correctness. Gating the process score by correctness preserves the answer objective and introduces graded preferences among correct trajectories. Human traces affect training only through this reward pathway and never enter the policy context. 

In Section~\ref{sec:human_process_data}, we describe how human interaction signals are collected and distilled into reference traces. In Section~\ref{sec:human_anchored_evaluation}, we introduce the human-anchored process judge. In Section~\ref{sec:answer_gated_reward}, we present the policy optimization with answer-gated reward.  Algorithms~\ref{alg:data} and~\ref{alg:haprl} summarize the data-construction and training procedures.

\begin{center}
\centering
\captionsetup{hypcap=false}
\includegraphics[width=0.94\linewidth]{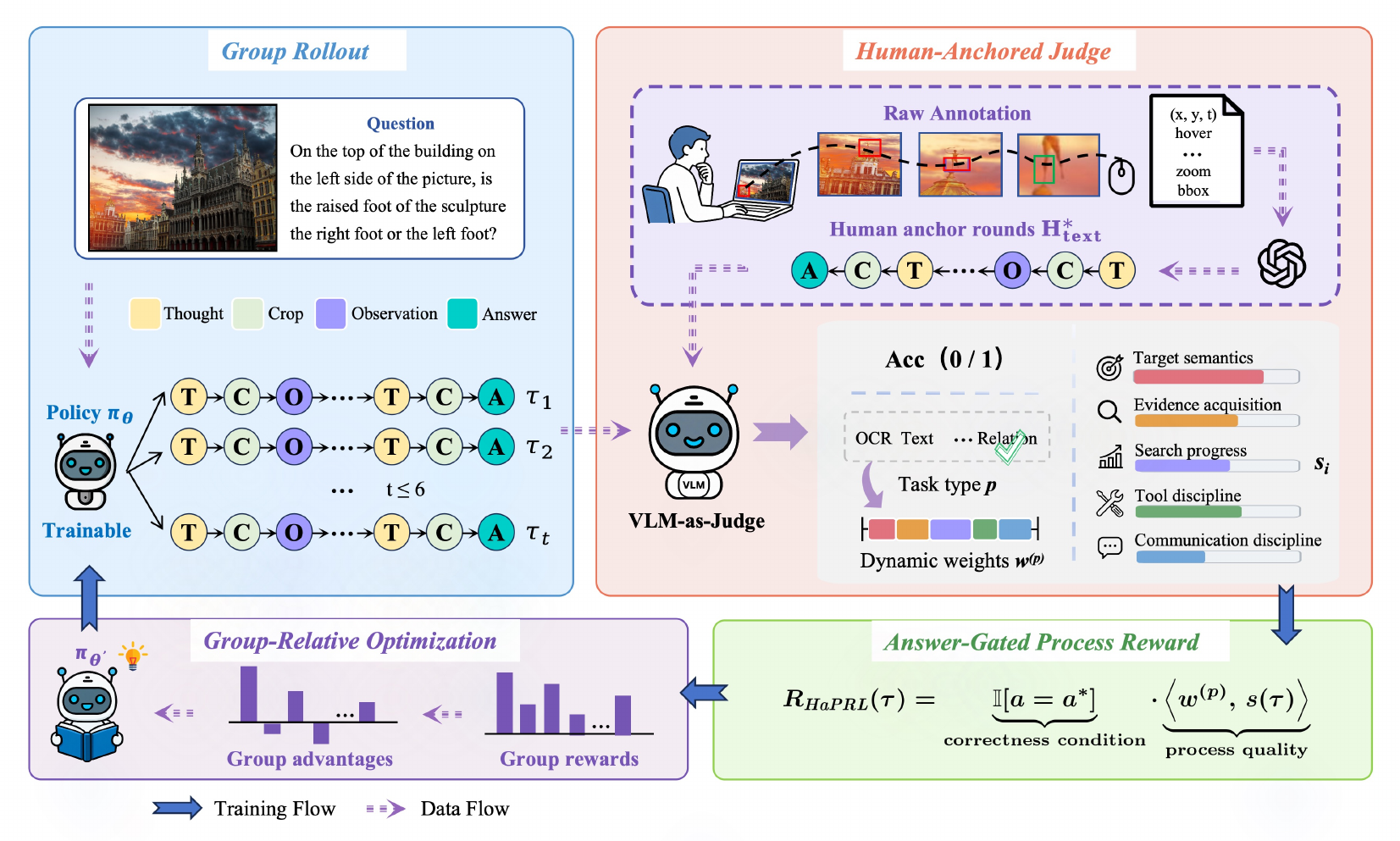}
\captionof{figure}{\textbf{\method{} converts human search behavior into graded rewards for visual-search training.} Human interaction logs are distilled into reference traces. A frozen vision-language judge then compares each sampled trajectory with its matched trace along five task-adaptive dimensions, while answer correctness gates the weighted process score. Finally, group-relative optimization uses the resulting rewards to update the policy model.}
\label{fig:method_pipeline}
\end{center}

\subsection{Human Process Data Collection}
\label{sec:human_process_data}

\paragraph{Interaction logging.} We build a browser-based platform and collect human search behavior for $1{,}104$ questions from the VisualProbe training set.
For each question, the human annotator searches the full-resolution image and uses the mouse to mark the region containing the supporting evidence.
The platform automatically records the interaction stream $E^*=(e_1,\ldots,e_M)$ throughout this process. We represent each event as $e_m=(\kappa_m,t_m,\mathbf{p}_m,v_m,\alpha_m,b_m)$, where $\kappa_m$ is the event type, $t_m$ is its timestamp, $\mathbf{p}_m$ is the pointer position, $v_m$ and $\alpha_m$ are its velocity and acceleration, and $b_m$ is an optional zoom or selection box. \emph{These measurements retain the temporal and spatial structure of the search}, including scan direction, dwell, redirection, and progressive refinement.

\paragraph{Trace distillation.} Raw interaction streams are long, redundant, and misaligned with the discrete rounds of an agent trajectory. An offline distillation stage converts $E^*$ into an ordered textual trace $H^*=(r_1^*,\ldots,r_L^*)$. Each round $r_l^*$ records the inspected region, its source view, the normalized crop when available, and the evidence revealed at that stage. Consecutive pointer samples that do not change the inspected region are suppressed, while meaningful dwell, redirection, and coarse-to-fine refinement are retained. The resulting training example is $z=(I_0,q,a^*,E^*,H^*)$, where $a^*$ denotes the reference answer. The policy observes only $(I_0,q)$. The distilled trace and annotation metadata are reserved for reward computation. 
Thus, $H^*$ identifies the evidence requirements and search difficulty of an instance without prescribing the exact actions that the policy must reproduce.

\subsection{Human-Anchored Process Judge}
\label{sec:human_anchored_evaluation}

\paragraph{Human-anchored assessment.} A frozen vision-language judge $J_\phi$ takes $(I_0,q,\tau,H^*)$ as input. The matched human trace anchors the evaluation to the evidence requirements and search difficulty of the specific instance without prescribing the annotator's exact actions. The judge returns a score vector $\mathbf{s}(\tau)\in[0,1]^5$ over five dimensions. (i) Target semantics measures whether the search follows the queried object, attribute, text, count, or relation. (ii) Evidence acquisition measures whether the observed views provide sufficient support for the answer. (iii) Search progress captures information gain across successive actions. (iv) Tool discipline evaluates the validity and granularity of crop operations. (v) Communication discipline evaluates whether the reasoning remains concise and grounded in visible evidence. More details are provided in Appendix~\ref{app:structured_rubric}.

\paragraph{Task-adaptive scoring.} Different visual-search tasks require different forms of evidence, so the five criteria should not contribute equally. The judge assigns each instance to a profile $p\in\{\text{OCR/text},\text{count/relation},\text{attribute},\text{general search}\}$ and applies the corresponding weights $\mathbf{w}^{(p)}$. OCR/text places greater emphasis on evidence legibility, count/relation on spatial coverage and cross-region consistency, and attribute on precise target grounding. General search adopts a more balanced weighting across the shared criteria. Given the vector $\mathbf{s}(\tau)$, we compute the process score as:
\begin{equation}
\rproc(\tau)=\left\langle\mathbf{w}^{(p)},\mathbf{s}(\tau)\right\rangle=\sum_{k=1}^{5}w_k^{(p)}s_k(\tau),\qquad \sum_{k=1}^{5}w_k^{(p)}=1.
\label{eq:process_score}
\end{equation}

\subsection{Answer-Gated Process Optimization}
\label{sec:answer_gated_reward}

\paragraph{Correctness-gated reward.} Process quality should \emph{distinguish successful trajectories without rewarding an incorrect answer}. A separate frozen answer judge $J_{\mathrm{ans}}$ compares the extracted answer $a(\tau)$ with the reference $a^*$:
\begin{equation}
\rout(\tau)=J_{\mathrm{ans}}\!\left(q,a(\tau),a^*\right)\in\{0,1\}.
\label{eq:outcome_reward}
\end{equation}
We invoke the process judge only when $\rout(\tau)=1$ and define the training reward as
\begin{equation}
R_{\text{\method}}(\tau)=\rout(\tau)\,\rproc(\tau).
\label{eq:haprl_reward}
\end{equation}
The gate assigns zero reward to incorrect trajectories and ranks correct trajectories by evidence-acquisition quality. In contrast, the outcome-based baseline uses $R_{\text{\orl}}(\tau)=\rout(\tau)$ and assigns the same reward to grounded, redundant, and lucky correct trajectories.

\paragraph{Group-relative optimization.}
\label{sec:group_relative_learning}
For each input $x=(I_0,q)$, we sample $G$ trajectories $\{\tau_i\}_{i=1}^{G}$ and standardize their rewards within the group:
\begin{equation}
A_i=\frac{R_i-\bar R}{\sqrt{G^{-1}\sum_{j=1}^{G}(R_j-\bar R)^2+\epsilon}},\qquad \bar R=G^{-1}\sum_{j=1}^{G}R_j.
\label{eq:group_advantage}
\end{equation}
An all-correct group receives a constant outcome reward and therefore yields zero relative advantage.
\method{} preserves within-group variation through $\rproc$, \emph{allowing successful trajectories to be distinguished by search quality}. All-wrong groups remain at zero because the correctness gate suppresses process credit. We optimize these advantages with the clipped GRPO objective \citep{shao2024deepseekmath}:
\begin{equation}
\mathcal{L}(\theta)=-\mathbb{E}_{i,t}\!\left[\min\!\left(\rho_{i,t}A_i,\operatorname{clip}(\rho_{i,t},1-\varepsilon,1+\varepsilon)A_i\right)\right]+\beta D_{\mathrm{KL}}(\pi_\theta\Vert\pi_{\mathrm{ref}}),
\label{eq:grpo_objective}
\end{equation}
where $\rho_{i,t}=\pi_\theta(y_{i,t}\mid x,y_{i,<t})/\pi_{\theta_{\mathrm{old}}}(y_{i,t}\mid x,y_{i,<t})$ is the token-level importance ratio, and $\pi_{\mathrm{ref}}$ is the reference policy used for KL regularization. 
At inference time, the learned policy interacts directly with the crop tool without judges or human traces.

\section{Experiments}
\label{sec:experiments}

\paragraph{Benchmarks and metrics.} We evaluate on VisualProbe Easy/Medium/Hard \citep{lai2025minio3}, V$^{\ast}$Bench QA \citep{wu2024vstar} and HR-Bench 4K/8K \citep{wang2025hrbench}. Each agent answers every question with one greedy rollout, capped at a maximum of six tool rounds. Base models are evaluated by answering in a single pass with no tool calls, so their rows report a starting point rather than a search policy. All models share that visual token budget, so no gap below comes from more pixels. Besides, V$^{\ast}$QA is graded open-ended by the judge rather than by option matching, which denies the policy the option-elimination shortcut. ``Avg.'' is the unweighted mean over the six sets. Search quality is scored separately on \emph{answer-correct} rollouts only, separating how an agent searched from how often it was right. See Appendix~\ref{app:benchmarks} for more details.

\paragraph{Training.} Four backbones span two families and scales: Qwen3-VL-4B/8B-Instruct \citep{bai2025qwen3vl} and LLaVA-OneVision-1.5-4B/8B \citep{an2025llavaonevision15}. None emits the grounding syntax reliably on its own, so each is cold-started on filtered Mini-o3 trajectories \citep{lai2025minio3}. Two RL conditions follow. \orl{} optimizes answer correctness alone, \method{} optimizes the answer-gated product.

\paragraph{Process annotation from Human.} Human annotations cover $1{,}104$ training questions from VisualProbe. Annotators worked in a browser interface logging hover and zoom motion, and each event stream is distilled into a per-round textual trace.

\paragraph{Hyper-parameters.} RL uses GRPO \citep{shao2024deepseekmath} in verl \citep{sheng2025hybridflow} with vLLM rollouts \citep{kwon2023vllm}: $8$ rollouts per prompt, prompt batch $48$, actor learning rate $5\times10^{-7}$, and KL coefficient $3\times10^{-3}$. The judge is a frozen Qwen3-VL-30B-A3B-Instruct. Both arms select checkpoints on the same held-out split. Remaining settings are in Appendices~\ref{app:coldstart}--\ref{app:judge}.

\subsection{Benchmarking \orl{}-based VLMs}
\label{subsec:main}

\paragraph{Comprehensive Improvements.} As shown in Table~\ref{tab:main}, \method{} wins all $24$ backbone-benchmark cells over \orl{} and the base model. Macro-average margins over matched \orl{} reach $\textbf{+30.3\%}$ on Qwen3-VL-4B ($44.26$ vs.\ $33.98$) and $\textbf{+15.3\%}$ on LLaVA-OneVision-1.5-8B ($47.80$ vs.\ $41.46$), with $\textbf{+13.8\%}$ and $\textbf{+5.8\%}$ on the other two.

\paragraph{SFT Trade-off Correction.} Cold-start SFT costs $3.00$ to $8.71$ points of accuracy and enables a policy that can call the crop tool. Outcome-only RL fails to repay that debt on half the backbones, leaving Qwen3-VL-4B $6.09$ points below its own base model ($33.98$ vs.\ $40.07$). \method{} clears the base model on all four, by $\textbf{+4.19}$ to $\textbf{+7.40}$ points. Qwen3-VL-4B on HR-Bench 4K is the cleanest case: the cold start gives up $6.62$ points ($48.88$ vs.\ $55.50$), \orl{} a further $1.75$, and \method{} recovers $7.62$ of them ($56.50$). The cold start is a wager that tool use is worth an accuracy deficit, and \emph{only the process reward collects on it}.

\begin{table}[t]
\centering
\caption{Accuracy comparison of generalist VLMs, SFT, +outcome-based RL, and +our method (\method{}) on VisualProbe Easy/Medium/Hard \citep{lai2025minio3}, V$^{\ast}$QA \citep{wu2024vstar}, HR-4K/8K \citep{wang2025hrbench}. All RL post-training methods are trained for three epochs. Avg.\ is the unweighted mean over the six sets, and the parenthesized value is the \method{} margin over the matched \orl{} control. Best per column of the same backbone in \textbf{bold}.}
\label{tab:main}
\small
\setlength{\tabcolsep}{3.6pt}
\begin{tabular}{lcccccccl}
\toprule
& \multicolumn{3}{c}{VisualProbe} & \multicolumn{3}{c}{High-resolution QA} & & \\
\cmidrule(lr){2-4}\cmidrule(lr){5-7}
Model & Easy & Medium & Hard & V$^{\ast}$QA & HR-4K & HR-8K & Avg. & \\
\midrule
\multicolumn{9}{l}{\textit{Qwen3-VL-4B-Instruct}}\\
\quad Base            & 36.88 & 16.42 & 13.21 & 67.16 & 55.50 & 51.25 & 40.07 & \\
\quad +SFT            & 27.66 & 14.55 &  6.60 & 46.07 & 48.88 & 44.38 & 31.36 & \\
\quad +SFT+\orl{} & 29.08 & 17.91 &  8.49 & 58.64 & 47.13 & 42.63 & 33.98 & \\
\rowcolor{basegrey} \quad +SFT+\method{} & \textbf{41.13} & \textbf{26.49} & \textbf{20.75} & \textbf{68.06} & \textbf{56.50} & \textbf{52.63} & \textbf{44.26} & \gain{(+10.28)}\\
\midrule
\multicolumn{9}{l}{\textit{Qwen3-VL-8B-Instruct}}\\
\quad Base            & 34.04 & 18.66 & 13.21 & 71.59 & 64.25 & 56.38 & 43.02 & \\
\quad +SFT            & 26.95 & 21.64 & 19.81 & 45.03 & 54.75 & 45.50 & 35.61 & \\
\quad +SFT+\orl{} & 34.04 & 27.61 & 15.09 & 56.54 & 70.25 & 62.25 & 44.30 & \\
\rowcolor{basegrey} \quad +SFT+\method{} & \textbf{36.88} & \textbf{36.57} & \textbf{20.09} & \textbf{72.73} & \textbf{70.38} & \textbf{65.88} & \textbf{50.42} & \gain{(+6.12)}\\
\midrule
\multicolumn{9}{l}{\textit{LLaVA-OneVision-1.5-4B}}\\
\quad Base            & 34.75 & 11.94 & 13.21 & 60.21 & 62.75 & 58.13 & 40.17 & \\
\quad +SFT            & 26.95 & 18.66 &  9.43 & 47.12 & 64.13 & 56.75 & 37.17 & \\
\quad +SFT+\orl{} & 41.13 & 18.66 & 17.92 & 60.21 & 63.50 & 56.25 & 42.95 & \\
\rowcolor{basegrey} \quad +SFT+\method{} & \textbf{43.97} & \textbf{20.52} & \textbf{23.58} & \textbf{61.26} & \textbf{64.75} & \textbf{58.50} & \textbf{45.43} & \gain{(+2.48)}\\
\midrule
\multicolumn{9}{l}{\textit{LLaVA-OneVision-1.5-8B}}\\
\quad Base            & 41.13 & 11.19 & 16.04 & 64.40 & 62.63 & 57.00 & 42.07 & \\
\quad +SFT            & 31.21 & 18.66 & 10.38 & 53.40 & 60.38 & 59.25 & 38.88 & \\
\quad +SFT+\orl{} & 42.55 & 16.04 & 12.26 & 63.87 & 60.13 & 53.88 & 41.46 & \\
\rowcolor{basegrey} \quad +SFT+\method{} & \textbf{50.52} & \textbf{18.28} & \textbf{27.92} & \textbf{65.97} & \textbf{64.38} & \textbf{59.75} & \textbf{47.80} & \gain{(+6.34)}\\
\bottomrule
\end{tabular}
\end{table}

\paragraph{Hard-search Gains.} A binary outcome reward takes two values inside a sampled group, so its advantage is zero among the correct rollouts and zero among the incorrect ones. The only gradient it supplies separates right answers from wrong ones. \emph{\method{} gates a rubric score by correctness leaves the incorrect set at zero and turns the correct set into a ranking, placing gradient exactly where outcome supervision has none.} Specifically, against the own cold start, \orl{} gains $+11.91$ on V$^{\ast}$QA and $+8.51$ on VisualProbe Easy but only $+1.68$ and $+1.89$ on Medium and Hard, the splits demanding the longest search. \method{} inverts that profile, adding $\textbf{+71.8\%}$ on VisualProbe Hard on average and peaking at $\textbf{+144.4\%}$ on Qwen3-VL-4B ($20.75$ vs.\ $8.49$) and $\textbf{+127.7\%}$ on LLaVA-OneVision-1.5-8B ($27.92$ vs.\ $12.26$). \emph{That is, an outcome reward can only sharpen a decision the answer prior already informs. A process reward orders the correct rollouts among themselves, and that ordering is where search discipline lives.}

\subsection{Search Quality under Different Settings}
\label{subsec:process_quality}

\begin{table}[!ht]
\centering
\caption{The process score (\%) on answer-correct rollouts of different methods on Qwen3-VL-4B; $\Delta$  against the cold start. The best results are in \textbf{bold}. Outcome-RL raises answer accuracy while leaving some search quality flat, whereas \method{} maintains a positive gain on every split.}
\label{tab:process_quality}
\small
\setlength{\tabcolsep}{5pt}
\begin{tabular}{lccccccc}
\toprule
& \multicolumn{3}{c}{VisualProbe} & \multicolumn{3}{c}{High-resolution QA} & \\
\cmidrule(lr){2-4}\cmidrule(lr){5-7}
Model & Easy & Medium & Hard & V$^{\ast}$QA & HR-4K & HR-8K & Avg. \\
\midrule
+SFT                  & 84.90 & 86.05 & 71.54 & 80.34 & 86.49 & 82.64 & 81.99 \\
+SFT+\orl{}     & 83.55 & 86.92 & 65.56 & 83.22 & 87.24 & 85.96 & 82.08 \\
$\Delta$              & \loss{$-$1.35} & \gain{+0.87} & \loss{$-$5.98} & \gain{+2.88} & \gain{+0.75} & \gain{+3.32} & \neut{+0.08} \\
\midrule
\rowcolor{basegrey}+SFT+\method{}     & \textbf{86.71} & \textbf{87.27} & \textbf{75.28} & \textbf{83.24} & \textbf{88.66} & \textbf{87.47} & \textbf{84.77} \\
$\Delta$              & \gain{+1.81} & \gain{+1.22} & \gain{+3.74} & \gain{+2.90} & \gain{+2.17} & \gain{+4.83} & \gain{+2.78} \\
\bottomrule
\end{tabular}
\end{table}

Higher accuracy does not imply better search. Table~\ref{tab:process_quality} scores answer-correct rollouts only, and \orl{} moves search quality by $+0.08$ points over its cold start ($82.08$ vs.\ $81.99$) while falling on two of six benchmarks. \method{} adds $\textbf{+2.78}$ ($84.77$) and improves all six.

VisualProbe Hard exposes the failure directly. \orl{} lifts accuracy there by $+28.6\%$ ($8.49$ vs.\ $6.60$) while its process score drops $8.4\%$ ($65.56$ vs.\ $71.54$): the policy answers more hard questions correctly and searches worse doing it. \method{} recovers $\textbf{+14.8\%}$ over that control ($75.28$ vs.\ $65.56$). An outcome reward is not neutral toward search; it erodes it. Because \emph{it cannot tell an answer earned by search from one reached by random guessing, accuracy on the hardest split rises while search quality falls.} This is \emph{reward hacking} whose signature is invisible to the metric the field reports.

\subsection{Scaling Behavior}
\label{subsec:scaling}

\begin{table}[t]
\centering
\caption{Accuracy of three-epoch training at different data sizes on Qwen3-VL-4B. Accuracy rises monotonically with the number of process-annotated prompts. The best is \textbf{bold}.}
\label{tab:data_scaling}
\small
\setlength{\tabcolsep}{5pt}
\begin{tabular}{lccccccc}
\toprule
& \multicolumn{3}{c}{VisualProbe} & \multicolumn{3}{c}{High-resolution QA} & \\
\cmidrule(lr){2-4}\cmidrule(lr){5-7}
Annotated prompts & Easy & Medium & Hard & V$^{\ast}$QA & HR-4K & HR-8K & Avg. \\
\midrule
$200$   & 31.21 & 14.18 & 16.98 & 57.07 & 51.38 & 48.88 & 36.62 \\
$400$   & 32.70 & 15.19 & 17.92 & 56.97 & 52.88 & 48.12 & 37.30 \\
$600$   & 35.46 & 17.16 & 18.87 & 60.40 & 54.25 & 49.13 & 39.21 \\
$800$   & 37.99 & 20.19 & 19.21 & 63.40 & 55.88 & 50.94 & 41.27 \\
$1000$  & 40.15 & 24.34 & 20.55 & 65.69 & 56.12 & 51.68 & 43.09 \\
$1104$ (full) & \textbf{41.13} & \textbf{26.49} & \textbf{20.75} & \textbf{68.06} & \textbf{56.50} & \textbf{52.63} & \textbf{44.26} \\
\midrule
$\Delta$ ($200\rightarrow1104$) & \gain{+9.92} & \gain{+12.31} & \gain{+3.77} & \gain{+10.99} & \gain{+5.12} & \gain{+3.75} & \gain{+7.64} \\
\bottomrule
\end{tabular}
\end{table}

\begin{figure}[t]
\begin{center}
\includegraphics[width=\linewidth]{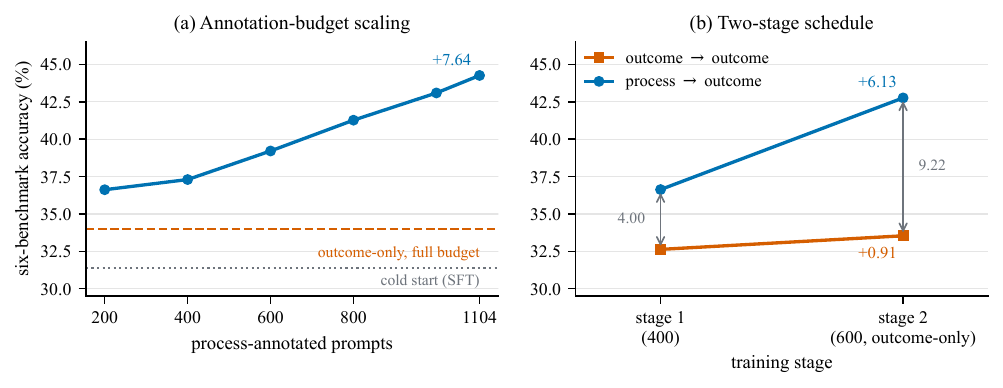}
\end{center}
\caption{\method{} \textbf{scales with the amount of process-annotated data}, and \textbf{injecting human-aligned process data in the early stage enables stronger scaling} even when later training uses only the outcome signal. (a) Accuracy grows monotonically with the process-annotation budget. 1/5 annotated data already beat Outcome-RL trained on all of the them. (b) A second, outcome-only stage returns +0.91 points to an Outcome-RL-trained policy and +6.13 points to a \method{}-trained one (\textbf{6.7×}).}
\label{fig:scaling}
\end{figure}

\paragraph{Accuracy scales with annotation volume.} Accuracy rises at every budget in Table~\ref{tab:data_scaling} and Figure~\ref{fig:scaling}(a), from $36.62$ at $200$ annotated prompts to $44.26$ at $1{,}104$ ($\textbf{+20.9\%}$), with no flattening at the largest budget we collected. Annotating $18\%$ of the prompts already beats \orl{} trained on all of them ($36.62$ vs.\ $33.98$). Returns concentrate on VisualProbe Medium ($\textbf{+12.31}$) and V$^{\ast}$QA ($\textbf{+10.99}$) and thin out on VisualProbe Hard ($\textbf{+3.77}$) and HR-Bench 8K ($\textbf{+3.75}$). \emph{Thus, process annotation is meaningful and scales the accuracy.}

\paragraph{Early process supervision scales better.} Table~\ref{tab:two_stage} and Figure~\ref{fig:scaling}(b) hold stage~2 identical across two arms and vary only stage~1. The same $600$-prompt outcome-only continuation returns $+2.8\%$ to the \orl{}-trained policy ($33.54$ vs.\ $32.63$) and $+16.7\%$ to the \method{}-trained one ($42.76$ vs.\ $36.63$), a $6.7\times$ difference produced by the initialization alone. The gap between arms widens from $+4.00$ to $+9.22$ points, uniformly across all six benchmarks ($+7.73$ to $+11.06$).

\begin{table}[t]
\centering
\caption{Accuracy of one-epoch training on Qwen3-VL-4B. Stage 1 trains on 400 data under either reward (outcome-based vs. HaPRL); stage 2 continues under the same outcome reward on another 600 data. The best is \textbf{bold}.}

\label{tab:two_stage}
\small
\setlength{\tabcolsep}{4pt}
\begin{tabular}{lccccccc}
\toprule
& \multicolumn{3}{c}{VisualProbe} & \multicolumn{3}{c}{High-resolution QA} & \\
\cmidrule(lr){2-4}\cmidrule(lr){5-7}
Training schedule & Easy & Medium & Hard & V$^{\ast}$QA & HR-4K & HR-8K & Avg. \\
\midrule
\multicolumn{8}{l}{\textit{Stage 1 only ($400$ prompts)}}\\
\quad \orl{} & 28.35 & 14.94 & 11.21 & 54.84 & 46.12 & 40.30 & 32.63 \\
\quad \method{} & 32.70 & 15.19 & 17.92 & 56.97 & 52.88 & 44.12 & 36.63 \\
\quad $\Delta$ & \gain{+4.35} & \gain{+0.25} & \gain{+6.71} & \gain{+2.13} & \gain{+6.76} & \gain{+3.82} & \gain{+4.00} \\
\midrule
\multicolumn{8}{l}{\textit{Stage 1 $\rightarrow$ outcome-only stage 2 ($600$ prompts)}}\\
\quad \orl{} $\rightarrow$ \orl{} & 28.68 & 16.85 & 9.34 & 57.84 & 47.03 & 41.51 & 33.54 \\
\rowcolor{basegrey} \quad \method{} $\rightarrow$ \orl{} & \textbf{39.74} & \textbf{24.58} & \textbf{19.36} & \textbf{66.42} & \textbf{55.21} & \textbf{51.27} & \textbf{42.76} \\
\quad $\Delta$ & \gain{+11.06} & \gain{+7.73} & \gain{+10.02} & \gain{+8.58} & \gain{+8.18} & \gain{+9.76} & \gain{+9.22} \\
\bottomrule
\end{tabular}
\end{table}

That is, \emph{process data can first establish correct search behavior, and a later outcome signal amplifies what is already working.} Outcome supervision used throughout has no such foundation to build on. It accepts rollouts that reach the right answer through a faulty search, and a policy trained on those trajectories has little real signal left to learn from.

\subsection{Ablation Study}
\label{subsec:ablation}
We further ablate the human reference trace in three ways on Qwen3-VL-4B, as shown in Table~\ref{tab:ablation}.

\begin{table}[t]
\centering
\caption{Ablation study of different training variants. Six-benchmark average accuracy of three-epoch training on Qwen3-VL-4B; $\Delta$ against full \method{}. Shuffled re-pairs each reference trace with another question’s trajectory, keeping the label distribution intact. Judge-only removes the trace from the judge prompt and leaves the rubric and task weighting intact. Noise perturbs 10\% of the recorded dwell and hover events.}
\label{tab:ablation}
\small
\setlength{\tabcolsep}{6pt}
\begin{tabular}{lcccc}
\toprule
Variant & Human trace & Correct pairing & Avg. & $\Delta$ \\
\midrule
Shuffled process labels & \checkmark & & 28.63 & \loss{$-$15.63} \\
Judge-only process      &            & & 36.41 & \loss{$-$7.85} \\
$+$ $10\%$ trace noise  & \checkmark & \checkmark & 41.74 & \loss{$-$2.52} \\
\textbf{\method{} (full)} & \checkmark & \checkmark & \textbf{44.26} & \neut{--} \\
\bottomrule
\end{tabular}
\end{table}

\paragraph{Shuffled pairing.} Re-pairing each trace with another question's trajectory preserves the label distribution and destroys only the alignment, at a cost of $35.3\%$ ($28.63$ vs.\ $44.26$). Falling $5.35$ points below \orl{}, and $2.73$ below the cold start, is the informative part. A broken regularizer would decay toward the unregularized baseline, whereas a mis-paired trace pushes the policy toward search that was correct for a different image.

\paragraph{Judge without the human trace.} Dropping the human-annotated reference trace and letting the frozen judge score the same five-dimensional rubric on its own gives $36.41$, only $+2.43$ over \orl{}. Restoring the trace adds $\textbf{+7.85}$ points ($+21.6\%$, $44.26$ vs.\ $36.41$), which is $76\%$ of the full $+10.28$ margin. The rubric supplies the axes of judgement; the trace supplies where good search sits on those axes for this particular image. A judge denied the trace has to invent that reference from the question alone with its own knowledge. \emph{Thus, the human annotated data is important and carries the learning signal.}

\paragraph{Robustness to annotation noise.} Perturbing $10\%$ of the recorded dwell and hover events costs $5.7\%$ ($41.74$ vs.\ $44.26$) and still leaves $+7.76$ over \orl{}. What the reward takes from a trace is the order in which regions were worth visiting, and that ordering survives jitter in individual events.

Therefore, \emph{the process reward is a channel for human search behavior rather than a hand-designed prior. Its value tracks how faithfully a trace is paired with its trajectory.}

\section{Conclusion and Limitation}
\label{sec:conclusion}

Outcome-based reinforcement learning trains visual-search agents by scoring only the final answer, so a correct result reached through an incorrect search receives full credit and later optimization can scale along the wrong paths. Thus, \emph{we introduce \method{}, the first framework that reinforces the search process with human search behavior.} We collect $1{,}104$ human traces with fine-grained interaction signals and distill them into round-by-round descriptions of inspected regions and discovered evidence. A frozen judge then scores each rollout under task-adaptive weights, anchored on the matched human trace, and gates that score by answer correctness. Across all benchmarks, \method{} consistently outperforms matched outcome-based RL in both accuracy and search quality, and exhibits favorable scaling trends. These results have important implications for the training of foundation models and for expansion to more other domains.

\paragraph{Limitation \& Future Work.} \method{} offers a new path and training paradigm for future research. \emph{A natural next step is to extend this human-process alignment recipe to more domains}, especially those in which human annotation signals can play a central role, such as 3D rigging, GUI agents, etc..

\section*{AI use statement}
In this work, we used generative AI tools to polish manuscript wording after an author-written draft, and to assist with some training/inference codes.
All AI-assisted text and code were manually verified by the authors.
We take responsibility for the final content of this work.

\subsection*{Ethics statement}
This study strictly adheres to the ICLR Code of Ethics. The datasets utilized in our experiments are publicly available, fully anonymized, and do not involve human subjects, privacy infringement, or harmful discrimination concerns.

\subsection*{Reproducibility statement}
Implementation details and hyperparameter configurations are provided in the
paper and appendix. Training and evaluation code, together with the released
annotations, is available in the \href{https://github.com/zhangquanchen/HAPRL}{public HaPRL repository}.

\bibliography{iclr2027_conference}
\bibliographystyle{iclr2027_conference}

\newpage
\appendix
\section{Additional Method Details}
\label{app:method_details}

\subsection{Algorithms}
\label{app:algorithms}

Algorithm~\ref{alg:data} constructs the human process traces used as reward anchors. Algorithm~\ref{alg:haprl} is the subsequent training loop: sampled trajectories are scored by a frozen answer judge, process credit is assigned only when the answer is correct, and GRPO updates the policy from group-relative advantages. Human traces and both judges are used only in this training loop; inference retains the policy and the crop tool.

\begin{algorithm}[h]
\caption{Human process data construction}
\label{alg:data}
\begin{algorithmic}[1]
\Require VisualProbe training questions $\{(I_0,q,a^*)\}$
\Ensure Dataset $\mathcal{D}=\{(I_0,q,a^*,E^*,H^*)\}$
\State $\mathcal{D}\leftarrow\emptyset$
\For{each question $(I_0,q,a^*)$}
    \State Annotator searches $I_0$ and marks the region containing the supporting evidence
    \State Log the interaction stream $E^*=(e_1,\ldots,e_M)$, $e_m=(\kappa_m,t_m,\mathbf{p}_m,v_m,\alpha_m,b_m)$
    \State Distill $E^*$ into a round-by-round trace $H^*=(r_1^*,\ldots,r_L^*)$
    \State Suppress pointer samples that do not change the inspected region
    \State Retain dwell, redirection, and coarse-to-fine refinement
    \State $\mathcal{D}\leftarrow\mathcal{D}\cup\{(I_0,q,a^*,E^*,H^*)\}$
\EndFor
\State \Return $\mathcal{D}$ \Comment{the policy observes only $(I_0,q)$}
\end{algorithmic}
\end{algorithm}

\begin{algorithm}[h]
\caption{\method{} training with answer-gated process reward}
\label{alg:haprl}
\begin{algorithmic}[1]
\Require Policy $\pi_\theta$, frozen judges $J_{\mathrm{ans}}$ and $J_\phi$, dataset $\mathcal{D}$, group size $G$
\For{each training prompt $x=(I_0,q)$ with matched trace $H^*$ and answer $a^*$}
    \State Sample $G$ trajectories $\{\tau_i\}_{i=1}^{G}$ from $\pi_\theta$ with the crop tool
    \For{$i=1$ to $G$}
        \State $\rout(\tau_i)\leftarrow J_{\mathrm{ans}}(q,a(\tau_i),a^*)$
        \If{$\rout(\tau_i)=0$}
            \State $R_i\leftarrow 0$
        \Else
            \State $\mathbf{s}(\tau_i),p\leftarrow J_\phi(I_0,q,\tau_i,H^*)$
            \State Clip $\mathbf{s}(\tau_i)$ to $[0,1]$ and normalize $\mathbf{w}^{(p)}$
            \State $\rproc(\tau_i)\leftarrow\langle\mathbf{w}^{(p)},\mathbf{s}(\tau_i)\rangle$
            \State $R_i\leftarrow\rout(\tau_i)\,\rproc(\tau_i)$
        \EndIf
    \EndFor
    \State Compute group-relative advantages $A_i$ from $\{R_i\}_{i=1}^{G}$
    \State Update $\theta$ with clipped GRPO and KL regularization toward $\pi_{\mathrm{ref}}$
\EndFor
\end{algorithmic}
\end{algorithm}

\subsection{Human-Trace Representation}
\label{app:human_trace_representation}

For each annotated example, the raw event stream contains timestamps, cursor positions, hover and zoom events, selected regions, and the final response. We distill it offline into an ordered textual trace. Each retained round states the region under inspection, the normalized crop when available, and the evidence revealed by that view. Low-level pointer motion that does not change the inspected region is suppressed. This produces a compact semantic anchor that can be compared with the model rollout without requiring exact action or token alignment.

The process judge receives the original image, question, serialized model rollout, the matched human trace, and compact annotation metadata. The trace is identified by the same sample key as the rollout; it is never re-paired across examples except in the shuffled-reference ablation.

\subsection{Structured Process Rubric}
\label{app:structured_rubric}

The judge returns one score in $[0,1]$ for each of five dimensions. Target semantics measures whether the search follows the queried object, attribute, text, count, or relation. Evidence acquisition measures whether the observed views actually support the answer. Search progress rewards information gain and useful recovery while penalizing drift and repeated crops. Tool discipline covers valid source views, bounding boxes, and crop granularity. Communication discipline measures whether the reasoning is concise, consistent, and tied to visible evidence.

\begin{table}[htbp]
\centering
\caption{\textbf{Task-adaptive weights over the five process dimensions.}
Columns denote target semantics (Tar.), evidence acquisition (Evd.), search
progress (Prog.), tool discipline (Tool), and communication discipline (Com.).}
\label{tab:process_profile_weights}
\small
\setlength{\tabcolsep}{5pt}
\begin{tabular}{lccccc}
\toprule
Profile & Tar. & Evd. & Prog. & Tool & Com. \\
\midrule
General search   & .20 & .30 & .25 & .15 & .10 \\
OCR / text       & .15 & .40 & .20 & .15 & .10 \\
Count / relation & .20 & .35 & .25 & .10 & .10 \\
Attribute        & .25 & .30 & .20 & .15 & .10 \\
\bottomrule
\end{tabular}
\end{table}

The judge predicts the task profile from the question and annotations. Scores are clipped to $[0,1]$, profile weights are normalized to sum to one, and the weighted total is recomputed outside the model response. This prevents a malformed or internally inconsistent judge output from directly setting the reward.

\subsection{Reward Computation}
\label{app:reward_computation}

For every sampled trajectory, training applies the following sequence, also listed in Algorithm~\ref{alg:haprl}:
\begin{enumerate}[leftmargin=*,nosep]
\item Extract the terminal answer and obtain the binary semantic-match score $\rout$ from the answer judge.
\item If $\rout=0$, set $R_{\text{\method}}=0$ and skip process evaluation.
\item Otherwise, evaluate the rollout against its matched human trace, select the task profile, and compute $\rproc$ from the structured rubric.
\item Set $R_{\text{\method}}=\rout\rproc$, standardize rewards within the rollout group, and update the policy with GRPO.
\end{enumerate}

This ordering makes the gate semantic rather than cosmetic: fluent reasoning, valid tool syntax, or close imitation of the human trace cannot earn reward when the final answer is wrong. Conversely, answer-correct rollouts retain a continuous preference signal that outcome-only supervision discards.

\section{Experimental Details}
\label{app:details}

\subsection{Benchmarks}
\label{app:benchmarks}

VisualProbe \citep{lai2025minio3} splits questions by search difficulty, defined by target size and distractor density. V$^{\ast}$Bench QA \citep{wu2024vstar} scores the correct option text as an open-ended answer rather than as multiple choice, which removes the option-elimination shortcut. HR-Bench \citep{wang2025hrbench} pairs the same questions at 4K and 8K resolution, so the 8K split isolates the cost of searching a larger frame.

\paragraph{Visual token budget.} Every rollout, in training and in evaluation, may take at most six tool rounds and hold at most six images in context. Each image is rescaled into the range $4\times10^{4}$ to $10^{6}$ pixels. The budget is what makes the crop tool a real decision: magnifying a region costs one of six slots, so an agent that wastes rounds has fewer left for the target. All three agent conditions receive the identical budget, so no reported gap between them reflects one model seeing more pixels.

\paragraph{Metrics details.} Answer accuracy is a binary semantic match produced by the answer judge, which compares the string inside \texttt{\textless answer\textgreater} against the reference and accepts paraphrases and formatting differences. VisualProbe is natively short-answer and HR-Bench natively multiple choice. V$^{\ast}$Bench is also multiple choice, but we score its correct option text as a free-form answer, which stops the policy from recovering the answer by eliminating distractors instead of searching.

\subsection{Cold Start}
\label{app:coldstart}

Two filters reduce the $7{,}267$-trajectory Mini-o3 cold-start corpus \citep{lai2025minio3}. The first keeps trajectories with at most six assistant turns ($6{,}757$ left), matching the tool-round cap used in RL. The second drops trajectories dominated by recovery phrases or repeated long sentences ($6{,}057$ left), both of which survive supervised training and reappear as degenerate loops during rollout.

Fine-tuning updates the language model with the vision tower and multimodal projector frozen. We use learning rate $5\times10^{-6}$ for five epochs, sequence cutoff $32768$, and per-device batch $1$ with gradient accumulation $4$, under DeepSpeed ZeRO-3 offload \citep{rajbhandari2020zero} in LLaMA-Factory \citep{zheng2024llamafactory}. Images use the same pixel range as RL.

Cold-start loss is a poor predictor of multi-turn health: checkpoints with lower loss frequently produce repeated crops and recovery loops. We therefore select the RL initialization by rollout quality on $20$ held-out prompts, scoring answer rate, valid grounding rate, overlong rate, and mean response length.

\subsection{Reinforcement Learning}
\label{app:rl}

Table~\ref{tab:app_rl} gives the full RL configuration; the two conditions differ only in the optimization total. Both call the answer judge and the process judge and log all five rubric dimensions. Process quality therefore stays observable in the outcome-only arm, where it contributes nothing to the gradient.

\begin{table}[h]
\centering
\caption{RL configuration. The last row is the only difference between the two conditions.}
\label{tab:app_rl}
\small
\begin{tabular}{ll}
\toprule
Setting & Value \\
\midrule
Algorithm & GRPO \citep{shao2024deepseekmath} \\
Framework & verl \citep{sheng2025hybridflow}, vLLM rollout \citep{kwon2023vllm} \\
Rollouts per prompt & 8 \\
Prompt batch size & 48 \\
PPO mini / micro batch & 8 / 1 per GPU \\
Actor learning rate & $5\times10^{-7}$ \\
KL loss coefficient & $3\times10^{-3}$ (low-variance estimator) \\
Rollout temperature / top-$p$ & 1.0 / 1.0 \\
Max prompt / response tokens & 8192 / 8192 \\
Max tool rounds & 6 \\
Max images per context & 6 \\
Image pixel range & $4\times10^{4}$ to $10^{6}$ \\
Hardware & 8 $\times$ H100 or 8 $\times$ H20 \\
\bottomrule
\end{tabular}
\end{table}

\subsection{Judge}
\label{app:judge}

Both judges use a frozen Qwen3-VL-30B-A3B-Instruct served behind an OpenAI-compatible endpoint. The process judge is called at temperature $0$ with structured JSON output. It receives the original image, the full multi-turn rollout rendered as text, and the human reference trace for the same question. Three call outcomes are logged: \texttt{vision} when the image is accepted, \texttt{text\_fallback} when the image call fails and the judge is retried on text alone, and \texttt{failed} after three attempts. We serialize judge-heavy jobs and halt training when the fallback rate grows, since a degraded judge silently flattens the process reward.

The answer judge calls the same model through a separate text-only prompt  and returns a binary semantic-correctness score. It is shared by both training conditions and all evaluations.

\subsection{Process Annotation}
\label{app:annotation}

Annotators answered $1{,}104$ VisualProbe training questions in a browser interface over the full-resolution image. The interface logs a timestamped event stream, with event types covering canvas entry, hover movement, zoom start, and zoom motion, together with the cursor position and the region under inspection. Annotators also record the bounding boxes they settled on as the evidence for their answer.

Raw event streams never reach the judge. Each stream is distilled into a short per-round textual trace naming the region the annotator inspected and what they found there. That distilled trace is the human reference the judge reads. Annotation is collected once per question and reused across every run in the paper, and it never enters the policy's context at training or inference time.

\paragraph{Annotation cost.} A single annotator produced the whole collection. One question takes about $1.5$ minutes on average, so the $1{,}104$ questions amount to roughly $27.6$ person-hours. Because each trace is collected once and reused by every run reported in this paper, this cost is paid once for the entire study rather than per training run or per backbone. Appendix~\ref{app:cheap_ref} weighs it against cheaper reward references, including one that needs only the final evidence box and takes about ten seconds per question.

\paragraph{Ablation construction.} The shuffled-label variant permutes the mapping from questions to reference traces, so each rollout is judged against a trace collected for a different question. The label distribution is unchanged and only the correspondence is destroyed. The noise variant perturbs $10\%$ of the recorded dwell and hover events in each stream before distillation. The judge-only variant removes the reference trace from the judge prompt and leaves the rest of it, including the rubric and the task-profile weighting, intact.

\section{Additional Experiments}
\label{app:additional}

Three experiments separate the contribution of the human trace from three things that could be mistaken for it, i.e., (i) the contribution of any evidence specification, (ii) the contribution of a better cold start, and (iii) seed variance.

\subsection{Cheaper Reward References}
\label{app:cheap_ref}

Anchoring the process reward to an ordered human trace costs $27.6$ person-hours (Appendix~\ref{app:annotation}). Three cheaper references can occupy the same slot in the reward. A \emph{hand-designed tool-use bonus} pays a fixed amount for every crop that is not a near-duplicate of an earlier one, and needs neither annotation nor a judge call. A \emph{final evidence box} scores a rollout by how much of the box the annotator settled on its crops covered. A \emph{teacher-generated evidence path} asks a strong vision-language model, shown the ground-truth answer, to write the region sequence it would have inspected, and judge reads that path exactly as it reads a human one.

Table~\ref{tab:app_cheap_ref} holds training identical and varies only the reference. Every substitute improves on the outcome-only reward, so part of the gain follows from anchoring the reward to an evidence specification of any kind. The substitutes then separate by how much of the search each one describes. The tool-use bonus, which describes none of it, recovers $1.74$ points. The final box, which names where the evidence is, recovers $4.63$ points. The teacher path, which names an ordering as well, reaches $39.85$ with no human labour at all. The human trace adds a further $4.41$ points over the best substitute.

What separates the human trace from the teacher path is where the ordering comes from. \emph{A path written backwards from a known answer contains no rejected hypotheses}, so a rollout that inspects two plausible regions before converging is scored against a reference that went straight to the target. A human reference contains the same dead ends, and the judge can therefore distinguish productive exploration from redundant revisiting. 

\begin{table}[t]
\centering
\caption{Reward reference and what it specifies about the search. All arms train Qwen3-VL-4B for three epochs on the same $1{,}104$ prompts and differ only in what the process reward is anchored to. Avg. is the macro average over the six evaluation splits. The best is \textbf{bold}.}
\label{tab:app_cheap_ref}
\small
\setlength{\tabcolsep}{5pt}
\begin{tabular}{llccc}
\toprule
Reward reference & Specifies & Judge & Avg. & $\Delta$ vs.\ \method{} \\
\midrule
None (\orl{}) & --- & --- & 33.98 & \loss{$-10.28$} \\
Hand-designed tool-use bonus & crop novelty & --- & 35.72 & \loss{$-8.54$} \\
Rubric only, no reference & rubric dimensions & \checkmark & 36.41 & \loss{$-7.85$} \\
Final evidence box, coverage reward & location & --- & 38.61 & \loss{$-5.65$} \\
Teacher-generated evidence path & location, ordering & \checkmark & 39.85 & \loss{$-4.41$} \\
\rowcolor{basegrey} Human ordered trace (\method{}) & location, ordering, dead ends & \checkmark & \textbf{44.26} & --- \\
\bottomrule
\end{tabular}
\end{table}

\subsection{Cold-Start Dependence}
\label{app:cold_start}

Every backbone in Table~\ref{tab:main} loses accuracy after supervised fine-tuning, which leaves open whether \method{} improves visual search training or repairs that loss. We separate the two with a second cold start that does not degrade the backbone. Recipe~B keeps the unfiltered $7{,}267$-trajectory corpus, masks the loss on turns beyond the six-round cap instead of discarding those trajectories, and stops at two epochs rather than five. On Qwen3-VL-4B it lands at $39.84$, within $0.23$ of the $40.07$ the backbone reaches before any agent training.

Table~\ref{tab:app_cold_start} runs both reward conditions from each cold start. The advantage of \method{} over the outcome-only reward is $10.28$ points from Recipe~A and $5.87$ points from Recipe~B. Recovery from a weak initialization therefore accounts for $4.41$ of the $10.28$ points in the main table, and process supervision for the remaining $5.87$, measured from an initialization with nothing to recover. The two effects add rather than overlap: Recipe~B with \method{} reaches $49.02$, the highest score in this study and $4.76$ points above the best Recipe~A result.

\begin{table}[t]
\centering
\caption{Cold-start recipe crossed with reward condition on Qwen3-VL-4B. Recipe~A is the cold start used throughout the main text; Recipe~B keeps the unfiltered corpus, masks the loss beyond the six-round cap, and stops at two epochs. The backbone scores $40.07$ before any agent training. Entries are macro averages over the six evaluation splits.}
\label{tab:app_cold_start}
\small
\setlength{\tabcolsep}{2pt}
\begin{tabular}{lcccc}
\toprule
Cold-start recipe & SFT & $+$\orl{} & $+$\method{} & $\Delta$ (\method{} $-$ \orl{}) \\
\midrule
A: filtered corpus, five epochs (ours) & 31.36 & 33.98 & 44.26 & \gain{$+10.28$} \\
\rowcolor{basegrey} B: full corpus, over-turn masking, two epochs & 39.84 & 43.15 & \textbf{49.02} & \gain{$+5.87$} \\
\bottomrule
\end{tabular}
\end{table}

\subsection{Two-Stage Seed Variance}
\label{app:two_stage_seeds}

Table~\ref{tab:two_stage} reports one run per schedule, in which an outcome-only stage 2 adds $6.13$ points after a process-trained stage 1 and $0.91$ points after an outcome-only one. Table~\ref{tab:app_two_stage_seeds} repeats both schedules three times.

Across seeds the outcome-only path gains $0.88 \pm 0.06$ points in stage 2 and the process-initialized path gains $5.83 \pm 0.26$, a ratio of $6.6\times$. The largest spread in any row is $0.57$ points, an order of magnitude below the $4.95$-point difference between the two gains. Both schedules see the same $600$ prompts under the same reward in stage 2 and differ only in their stage-1 initialization, and the $3.94$-point gap they carry into stage 2 widens to $8.89$ by the end of it. The three seeds share the prompt ordering and the $400$/$600$ split, so these intervals cover optimization and rollout stochasticity rather than data partitioning.

\begin{table}[t]
\centering
\caption{Three seeds of the two-stage schedule of Table~\ref{tab:two_stage} on Qwen3-VL-4B. Seed~1 is the run reported in the main text. Entries are macro averages over the six evaluation splits; the last column is the stage-2 gain of each row over its own stage-1 checkpoint. Mean $\pm$ SD over three seeds.}
\label{tab:app_two_stage_seeds}
\small
\setlength{\tabcolsep}{5pt}
\begin{tabular}{lcccrc}
\toprule
Training schedule & Seed 1 & Seed 2 & Seed 3 & Mean $\pm$ SD & Stage-2 gain \\
\midrule
\multicolumn{6}{l}{\textit{Stage 1 only ($400$ prompts)}}\\
\quad \orl{}   & 32.63 & 32.05 & 32.55 & $32.41 \pm 0.31$ & --- \\
\quad \method{} & 36.63 & 35.94 & 36.48 & $36.35 \pm 0.36$ & --- \\
\midrule
\multicolumn{6}{l}{\textit{Stage 1 $\rightarrow$ outcome-only stage 2 ($600$ prompts)}}\\
\quad \orl{} $\rightarrow$ \orl{}   & 33.54 & 32.86 & 33.47 & $33.29 \pm 0.37$ & \gain{$+0.88 \pm 0.06$} \\
\rowcolor{basegrey} \quad \method{} $\rightarrow$ \orl{} & 42.76 & 41.62 & 42.16 & $\mathbf{42.18 \pm 0.57}$ & \gain{$+5.83 \pm 0.26$} \\
\bottomrule
\end{tabular}
\end{table}

\section{Qualitative Cases}
\label{app:cases}

The tables in Section~\ref{sec:experiments} report what changed; the six cases below show what the change looks like inside a rollout. Each figure holds the question fixed and renders the condensed trajectory of all three trained conditions from the same backbone: the cold start, \orl{}, and \method{}. Every round is shown with the reasoning that motivated it, the emitted crop, and the observation it returned, so the route to the answer can be read directly rather than inferred from a score.

These are six hand-picked examples chosen to span the task profiles of Section~\ref{sec:human_anchored_evaluation} (text recognition, spatial relation, and attribute recognition) and all three evaluation sources. They illustrate mechanisms already measured in aggregate and are not themselves evidence of frequency. Table~\ref{tab:app_cases} summarizes them.

\begin{table}[h]
\centering
\caption{The six qualitative cases. Rounds and crops are counted from the rendered trajectories in Figures~\ref{fig:case1}--\ref{fig:case6}. In every case \method{} reaches the correct answer with the shortest evidence path of the three conditions.}
\label{tab:app_cases}
\small
\setlength{\tabcolsep}{4pt}
\begin{tabular}{llllll}
\toprule
& & & \multicolumn{3}{c}{Rounds / crops and verdict} \\
\cmidrule(lr){4-6}
Fig. & Task profile & Source & +SFT & +\orl{} & +\method{} \\
\midrule
\ref{fig:case1} & OCR / text        & HR-Bench 4K-500        & 3 / 2 {\color{lossred}$\times$} & 6 / 5 {\color{gaingreen}$\checkmark$} & 4 / 3 {\color{gaingreen}$\checkmark$} \\
\ref{fig:case2} & OCR / text        & HR-Bench 8K-39         & 4 / 3 {\color{lossred}$\times$} & 7 / 6 {\color{lossred}$\times$}  & 3 / 2 {\color{gaingreen}$\checkmark$} \\
\ref{fig:case3} & OCR / text        & VisualProbe Easy-128   & 4 / 3 {\color{lossred}$\times$} & 7 / 6 {\color{lossred}$\times$}  & 2 / 1 {\color{gaingreen}$\checkmark$} \\
\ref{fig:case4} & OCR / relation    & VisualProbe Hard-54    & 7 / 6 {\color{lossred}$\times$} & 6 / 5 {\color{gaingreen}$\checkmark$} & 2 / 1 {\color{gaingreen}$\checkmark$} \\
\ref{fig:case5} & Attribute         & VisualProbe Medium-151 & 7 / 6 {\color{lossred}$\times$} & 7 / 6 {\color{lossred}$\times$}  & 3 / 2 {\color{gaingreen}$\checkmark$} \\
\ref{fig:case6} & Attribute         & V$^{\ast}$ attributes-58 & 4 / 3 {\color{lossred}$\times$} & 4 / 3 {\color{lossred}$\times$} & 2 / 1 {\color{gaingreen}$\checkmark$} \\
\midrule
\multicolumn{3}{l}{Mean rounds / crops} & 4.8 / 3.8 & 6.2 / 5.2 & \textbf{2.7 / 1.7} \\
\multicolumn{3}{l}{Correct} & 0 / 6 & 2 / 6 & \textbf{6 / 6} \\
\bottomrule
\end{tabular}
\end{table}

\begin{figure}[p]
\begin{center}
\includegraphics[width=\linewidth]{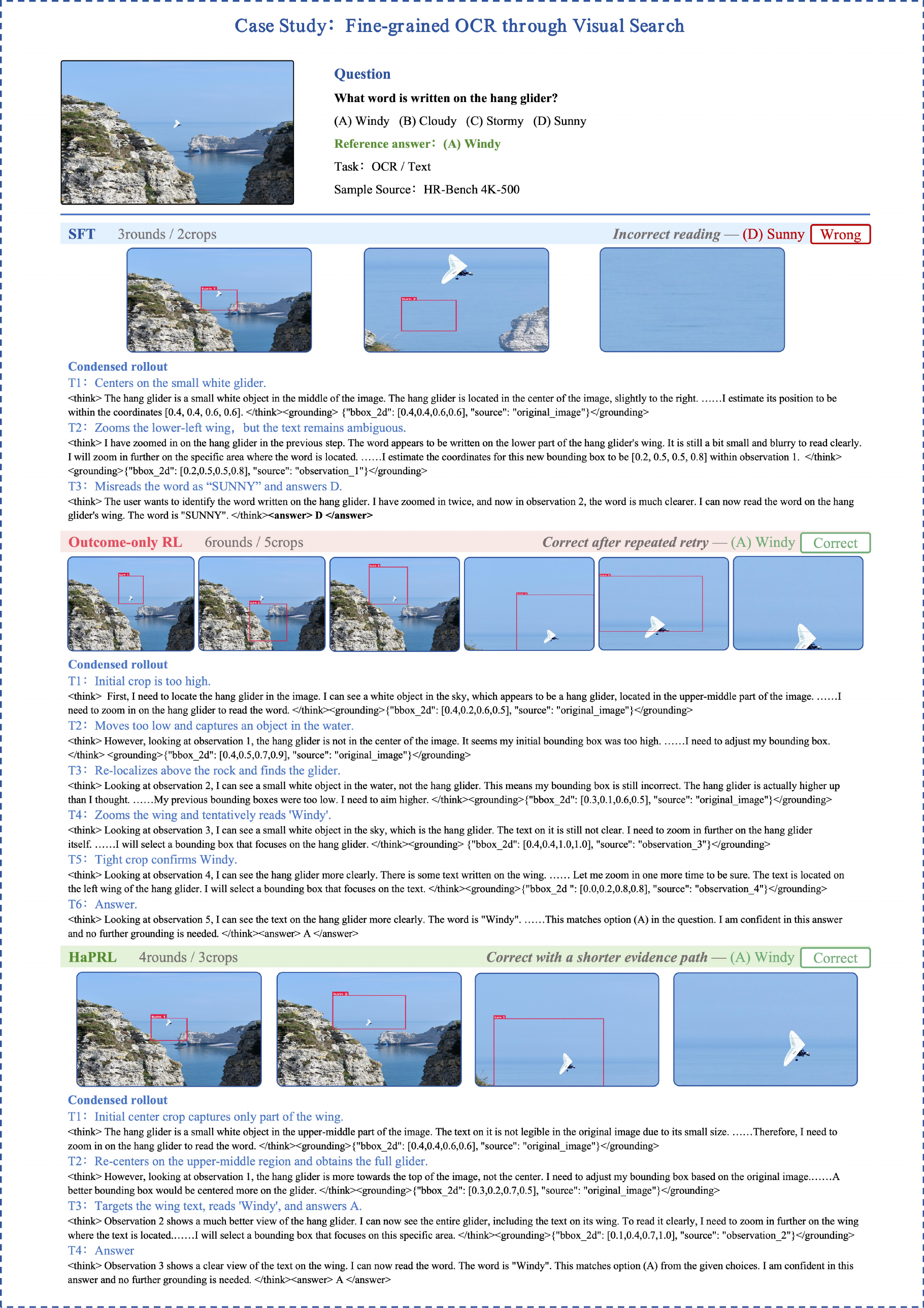}
\end{center}
\caption{\textbf{Fine-grained OCR through visual search} (HR-Bench 4K-500). The cold start zooms twice, misreads the word on the wing as ``SUNNY'', and answers the wrong option. \orl{} needs five crops, two of which land on water and rock, before reading the word correctly. \method{} re-centers once on the upper-middle region, targets the wing text, and answers in four rounds.}
\label{fig:case1}
\end{figure}

\begin{figure}[p]
\begin{center}
\includegraphics[width=\linewidth]{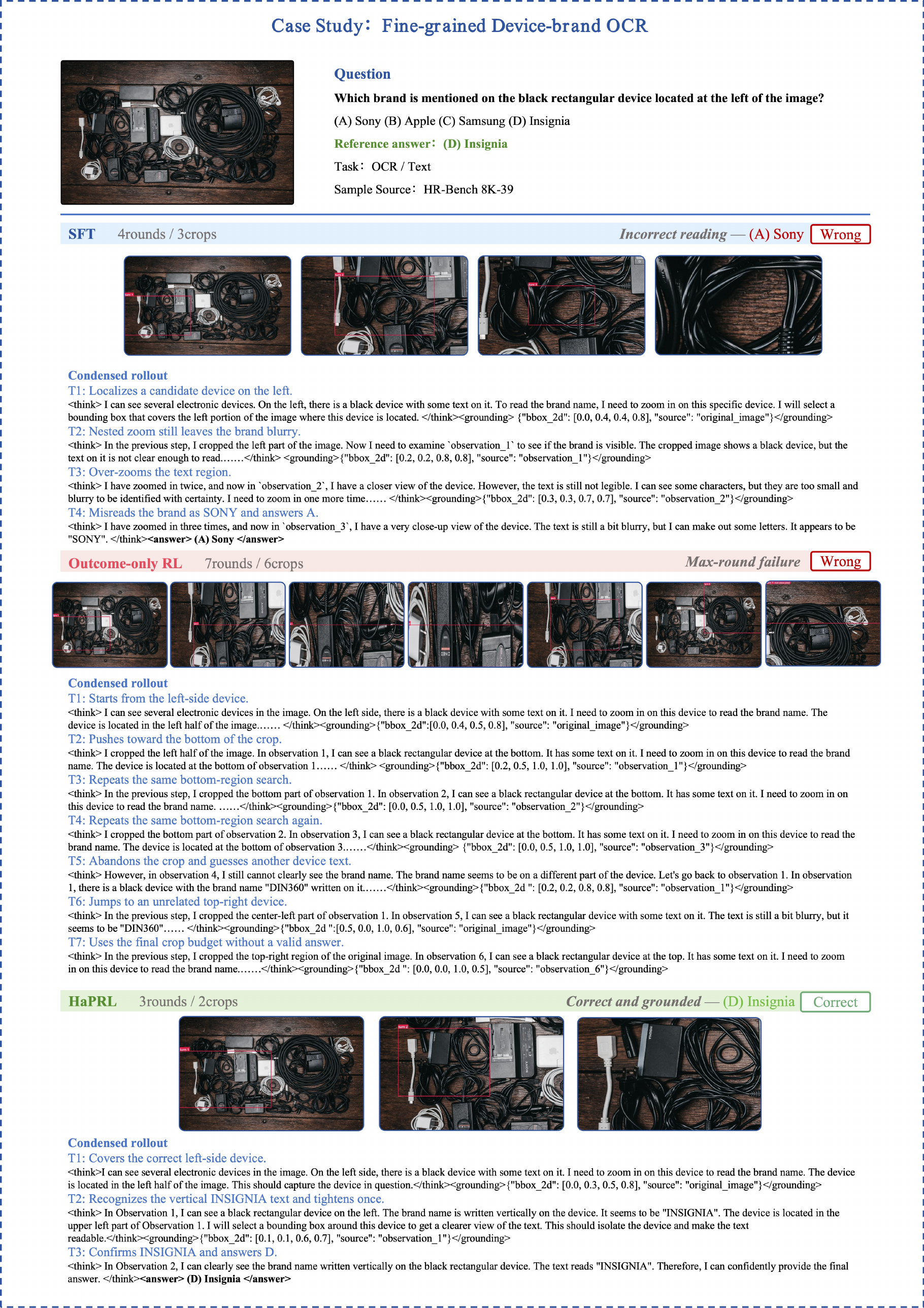}
\end{center}
\caption{\textbf{Fine-grained device-brand OCR} (HR-Bench 8K-39). The cold start over-zooms until the text is illegible and guesses ``Sony''. \orl{} repeats the same bottom-region crop three times, abandons it, jumps to an unrelated device, and exhausts its round budget without a valid answer. \method{} covers the left-side device, recognizes the vertical \textsc{insignia} text, and answers in three rounds.}
\label{fig:case2}
\end{figure}

\begin{figure}[p]
\begin{center}
\includegraphics[width=\linewidth]{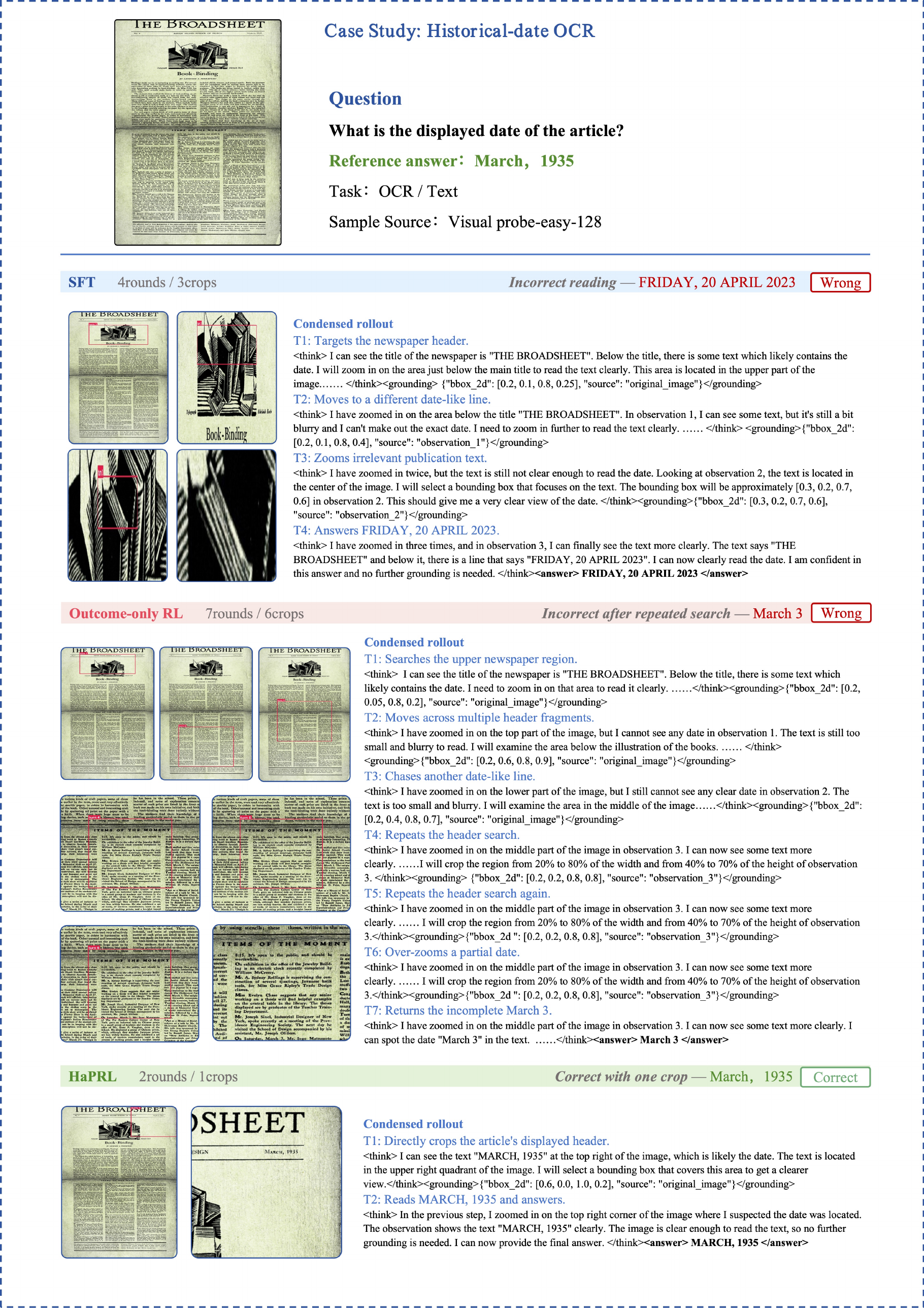}
\end{center}
\caption{\textbf{Historical-date OCR} (VisualProbe Easy-128). The cold start zooms into irrelevant publication text and reports a date that is not in the image. \orl{} chases several date-like lines across six crops and returns the truncated ``March 3''. \method{} crops the article's displayed header directly and reads ``March, 1935'' with one crop.}
\label{fig:case3}
\end{figure}

\begin{figure}[p]
\begin{center}
\includegraphics[width=\linewidth]{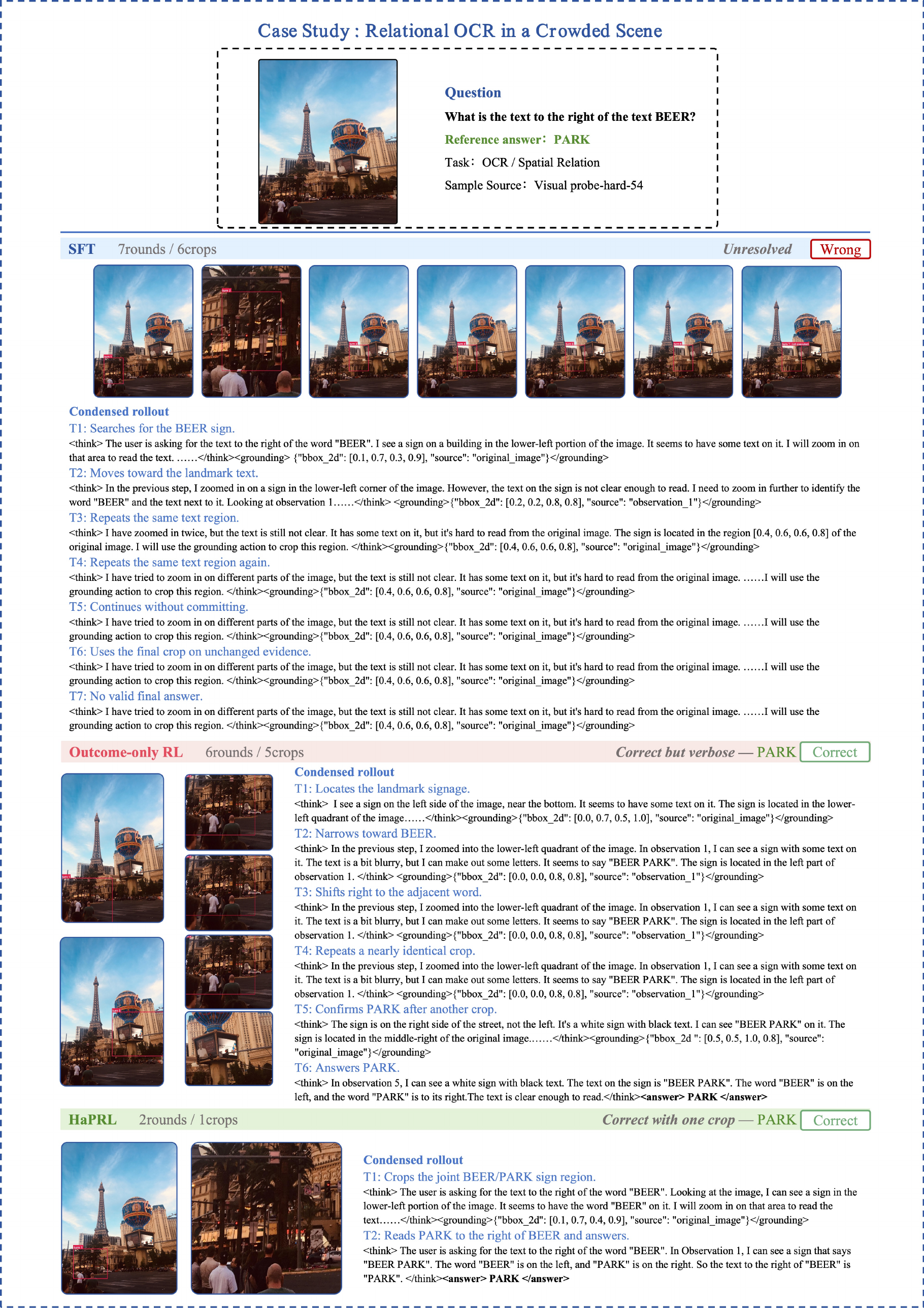}
\end{center}
\caption{\textbf{Relational OCR in a crowded scene} (VisualProbe Hard-54). The question asks for the text to the right of \textsc{beer}, so the two words must be legible in one view. The cold start re-crops the same unreadable region four times and returns no valid answer. \orl{} reaches the right answer but repeats a nearly identical crop and spends a further round confirming what it had already read. \method{} crops the joint sign region once, which makes the relation readable immediately.}
\label{fig:case4}
\end{figure}

\begin{figure}[p]
\begin{center}
\includegraphics[width=\linewidth]{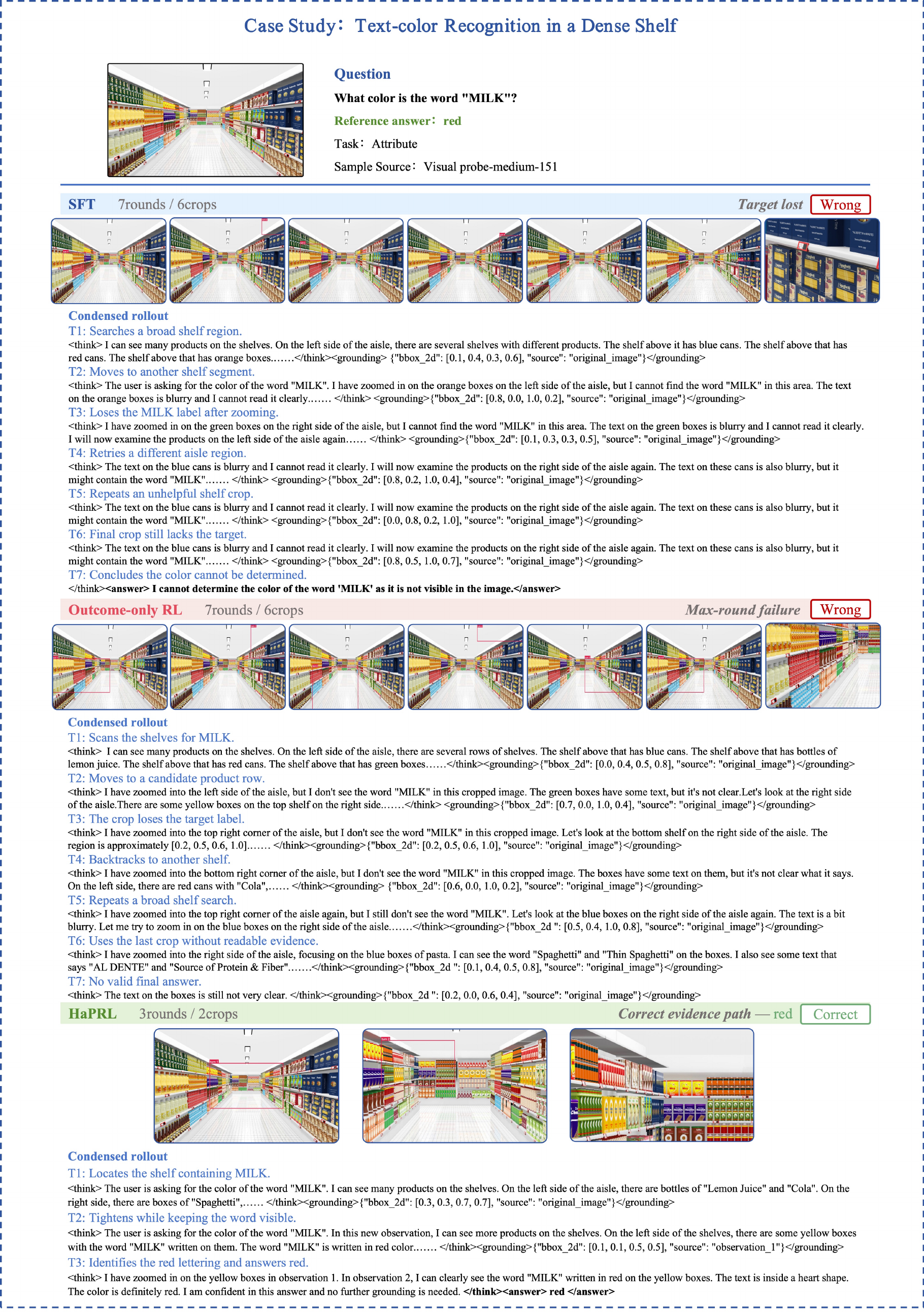}
\end{center}
\caption{\textbf{Text-color recognition on a dense shelf} (VisualProbe Medium-151). Both controls lose the target among repeated shelf crops and end without a usable answer, the cold start concluding the color cannot be determined. \method{} first locates the shelf holding the queried word, then tightens while keeping the word visible, and reads its color in three rounds.}
\label{fig:case5}
\end{figure}

\begin{figure}[p]
\begin{center}
\includegraphics[width=\linewidth]{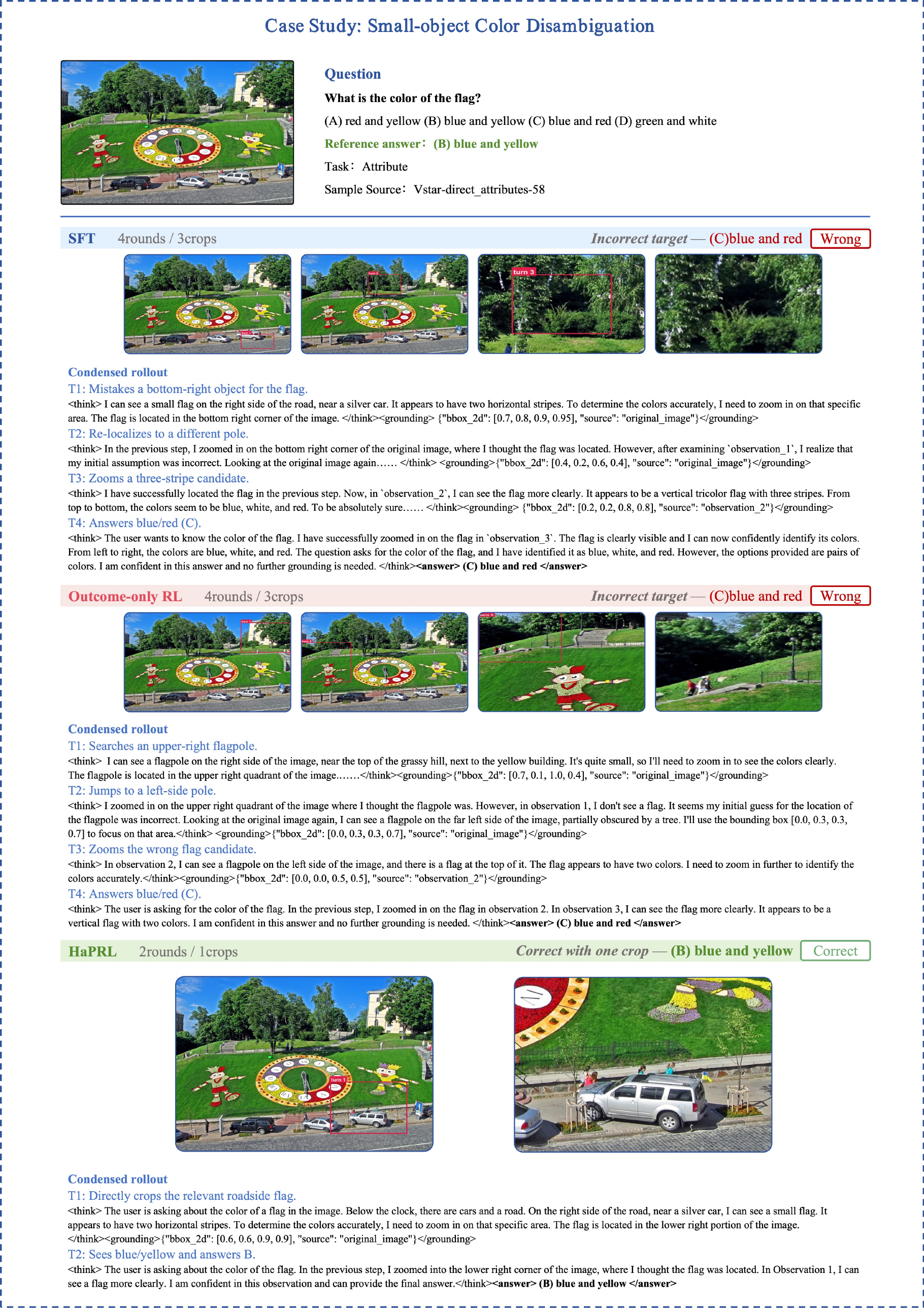}
\end{center}
\caption{\textbf{Small-object color disambiguation} (V$^{\ast}$ direct\_attributes-58). Both controls fix on a salient flagpole rather than the queried flag and report the colors of the wrong object, arriving at the same incorrect option by different routes. \method{} crops the roadside flag directly and answers in two rounds. A wrong target reached fluently is the failure the rubric's target-semantics dimension is meant to price.}
\label{fig:case6}
\end{figure}

\paragraph{What the outcome-only rollouts do with their extra rounds.} \orl{} uses the most rounds and the most crops of the three conditions and is correct twice. Both successes are the kind of rollout the argument of Section~\ref{sec:intro} predicts an outcome reward cannot penalize. In Figure~\ref{fig:case1} it overshoots the target twice, crops water and then rock, re-localizes, and arrives at the right word on the fifth crop. In Figure~\ref{fig:case4} it crops a nearly identical region twice in succession before confirming an answer it had already read. Neither trajectory is efficient and both receive full credit, because the only thing the reward can see is the string at the end. The four failures show where that indifference leads. In Figures~\ref{fig:case2} and \ref{fig:case5} the policy repeats the same bottom-region or shelf crop across three consecutive rounds, exhausts its six-round budget, and emits no valid answer. In Figure~\ref{fig:case3} it chases several date-like lines and returns the truncated ``March 3'' instead of ``March, 1935''. In Figure~\ref{fig:case6} it zooms to the wrong flagpole twice and reports the colors of a flag that is not the queried one, which the cold start does as well: \emph{repetition and distractor-chasing are exactly the behaviors a rubric can name and an outcome reward cannot.}

\paragraph{What the process-supervised rollouts do instead.} \method{} answers all six correctly using $2.7$ rounds and $1.7$ crops on average, fewer than either control. The pattern is consistent across profiles. In Figure~\ref{fig:case3} it crops the article header once and reads the date. In Figure~\ref{fig:case4} it crops the joint \textsc{beer}/\textsc{park} sign region once, which makes the spatial relation legible in a single view rather than requiring the two separate crops the other conditions attempt. In Figure~\ref{fig:case5} it locates the shelf containing the queried word and then tightens onto it while keeping the word visible, and in Figure~\ref{fig:case6} it crops the roadside flag directly instead of the salient flagpole above it. This is what the process-quality numbers of Table~\ref{tab:process_quality} measure: not that the answer is right, \emph{but that the view which justifies it was actually obtained, and obtained without spending rounds that reveal nothing new.}

\clearpage

\end{document}